\documentclass[letterpaper]{article} 
\usepackage{aaai23}  
\usepackage{times}  
\usepackage{helvet}  
\usepackage{courier}  
\usepackage[hyphens]{url}  
\usepackage{graphicx} 
\usepackage{natbib}  
\usepackage{caption} 
\usepackage{algorithm}
\usepackage{algorithmic}
\usepackage{amssymb}

\usepackage{newfloat}
\usepackage{listings}
\DeclareCaptionStyle{ruled}{labelfont=normalfont,labelsep=colon,strut=off} 
\floatstyle{ruled}
\newfloat{listing}{tb}{lst}{}
\floatname{listing}{Listing}
\newcommand{\ind}{\mathds{1}}
\newcommand{\bP}{\mathds{P}}
\newcommand{\bE}{\mathds{E}}
\newcommand{\Regret}{\operatorname{Regret}}

\newcommand{\ArmAt}{\hat{a}}
\newcommand{\defined}{\coloneqq}

\newcommand{\IsEqual}{=}

\newcommand{\KLUCBCF}{$\mathrm{kl}$-\textsc{UCB-CF}}
\newcommand{\SWKLUCBCF}{\textsc{SW-KLUCB-CF}}
\newcommand{\KLUCB}{$\mathrm{kl}$-\textsc{UCB}}

\newcommand{\cB}{\mathcal{B}}
\newcommand\given[1][]{\:#1\vert\:}
\newcommand{\MeanReAtIChanges}[2]{\mu_#1(#2)}
\newcommand{\optMeanAtIChanges}[1]{\mu_{*}(#1)}
\newcommand{\MeanFeAtIChanges}[2]{\lambda_#1(#2)}
\newcommand{\optArmAtIChanges}[1]{a_{*}(#1)}

\newcommand{\optArm}[1]{a_{*, #1}}
\newcommand{\optMean}[1]{\mu_{*, #1}}
\newcommand{\noChange}{\mathcal{T}}
\newcommand{\meanReward}[2]{\mu_{#1, #2}}
\newcommand{\meanFeedback}[2]{\lambda_{#1, #2}}
\newcommand{\timeOfChange}[1]{t_{#1}}
\newcommand{\numChanges}{L}

\newcommand{\RewardDist}[2]{\boldsymbol{\nu}_{#1}(#2)}
\newcommand{\FeedbackDist}[2]{\boldsymbol{\varsigma}_{#1}(#2)}
\newcommand{\windowsize}{w}

\newcounter{counterCorollary}

\newtheorem{myCorollary}[counterCorollary]{Corollary}
\newtheorem{myTheorem}{Theorem}
\newtheorem{myDefinition}{Definition}

\usepackage{amsmath}
\usepackage{kbordermatrix} 
\usepackage{mathtools}
\usepackage{dsfont}

\title{Local Differential Privacy for Sequential Decision Making in a Changing Environment}
\author{
  Pratik Gajane
}
\affiliations{
    Eindhoven University of Technology\\

    pratik.gajane@gmail.com
}

\begin{document}

\maketitle

\begin{abstract}
We study the problem of preserving privacy while still providing high utility in sequential decision making scenarios in a changing environment.  We consider abruptly changing environment: the environment remains constant during periods and it changes at unknown time instants. 
To formulate this problem, we propose a variant of multi-armed bandits called non-stationary stochastic corrupt bandits.
We construct an algorithm called \SWKLUCBCF \ and prove an upper bound on its utility using the performance measure of \textit{regret}. The proven regret upper bound for \SWKLUCBCF \ is near-optimal in the number of time steps and matches the best known bound for analogous problems in terms of the number of time steps and the number of changes. Moreover, we present a provably optimal mechanism which can guarantee the desired level of local differential privacy while providing high utility.
\end{abstract}

\section{Introduction}
Several practically relevant applications including recommender systems, Internet advertising have been formulated as sequential decision making problems using the framework of multi-armed bandits. The importance of privacy in such sequential decision making problems has been extensively discussed in the literature (see for example, 
\citet{DBLP:conf/nips/ThakurtaS13,DBLP:conf/uai/MishraT15,tossou:aaai2016}).

\textit{Differential privacy}, introduced by \citet{Dwork06calibratingnoise}, is one of the popular approaches to address such privacy concerns.  
In sequential decision making problems, algorithms providing differential privacy preserve data privacy by adding appropriate statistical noise to the data. 
\citet{Duchi:2014:PAL:2700084.2666468} extend this notion to \textit{local differential privacy} in which data remains private even from the algorithm. 
The main difference between global and local differential privacy is whether privacy is to be maintained from the algorithm or the (possibly unintended) recipient of the output of the algorithm. In global differential privacy, noise is added by the algorithm so the output does not reveal private information about the input. In local differential privacy, noise is added to the input of the algorithm so that privacy is maintained even from the algorithm. 

To understand the motivation for local differential privacy, let us consider the practical application of Internet advertising \footnote{We consider a simplistic scenario for illustrative purposes.}. An advertising system receives, as input, feedback from the users which may reveal private information about them. The advertising system employs a suitable learning algorithm and selects ads for the users tailored to the feedback given by them. These selected ads are then given to the advertisers as output. 
While using global differential privacy, privacy is maintained from the advertisers by ensuring that the output of the learning algorithms does not reveal information about the input (i.e., user information). Typically, advertising systems are established by leading social media networks, web browsers and other popular websites. \citet{DBLP:conf/icdm/Korolova10,kosinski2013private} show that it is possible to accurately predict a range of highly sensitive personal attributes including age, sexual orientation, relationship status, political and religious affiliation
using the feedback available to the advertising systems. Such possible breach of privacy necessitates us to protect personal user information not only from the advertisers but also from the advertising systems. Local differential privacy is able to achieve this objective unlike global differential privacy.

In this article, we propose to use \textit{low privacy} regime using local differential privacy.  
In low privacy regime, the noise added to the data is small and the aim of the privacy mechanism is to send as much information about data as allowed, but no more \citep{NIPS2014_5392}. This is in alignment with our dual goal of using privacy in recommendation systems or Internet advertising, and other similar applications: provide useful recommendations/ads to the users while respecting their privacy as much as possible.

We measure the utility of our proposed algorithm using \textit{regret} which is a measure of the total mistake cost (precise definitions will follow in the next Section). When rewards are bounded (as assumed in most works in the literature), the regret of any algorithm is trivially bounded linearly in the number of time steps $T$. An algorithm is said to be \textit{learning} if its regret is bounded sub-linearly in $T$.

\paragraph{Main Contributions}
\begin{enumerate}
    \item We propose non-stationary stochastic corrupt bandits, a novel formulation which aims to preserve local differential privacy while still providing high utility for sequential decision making in a non-stationary environment.
    \item We construct an algorithm called \SWKLUCBCF \ for the considered problem. 
    \item We prove an upper bound on the utility of \SWKLUCBCF \ in terms of its regret. This upper bound is near-optimal in terms of the number of time steps and matches the best known bound for analogous problems in terms of the number of time steps and the number of changes.
    \item We provide an optimal mechanism to achieve a desired level of local differential privacy while achieving high utility.
\end{enumerate}

This work is an extension of \citet{pmlr-v83-gajane18a} to non-stationary environments and reuses some of the concepts used there. However, it should be noted that the algorithms proposed in \citet{pmlr-v83-gajane18a} will not be able to solve the problem considered in this article. In fact, it is easy to construct non-stationary environments for which the algorithms proposed in \citet{pmlr-v83-gajane18a} (and all other differentially private algorithms designed for stationary environment) will suffer regret linear in the number of time steps $T$. On the other hand, the algorithm proposed in this article can guarantee regret sub-linear in $T$ in such scenarios. Furthermore, due to the changing environment and the use of a sliding window, the regret analysis in our article presents challenges not encountered in stationary settings. 

Our extension to non-stationary environments is practically relevant as the assumption of stationarity is sometimes unrealistic in real-world applications. Such an extension providing local differential privacy in non-stationary environments for the problem of data collection is given by \citet{NEURIPS2018_a0161022}. 
Our problem is different than \citet{NEURIPS2018_a0161022} as we study learning to make optimal sequential decisions in a non-stationary environment while providing local differential privacy. 
Note that a naive strategy of restarting an algorithm (designed for a stationary environment) after each change is not possible in the problem considered here as the time instants at which the changes occur are unknown.

\paragraph{Related Work}
In the context of sequential decision-making, global differential privacy has been studied in various settings including stochastic bandits \citep{DBLP:conf/uai/MishraT15, tossou:aaai2016}, adversarial bandits \citep{DBLP:conf/nips/ThakurtaS13, tossou:aaai2017a} and collaborative bandits \citep{10.1145/3383313.3412254}.  
In the context of sequential decision-making, local differential privacy has been considered in stochastic bandit setting \citep{pmlr-v83-gajane18a, pmlr-v151-tao22a}, contextual bandits \citep{NEURIPS2020_908c9a56}, collaborative bandits \citep{10.1145/3383313.3412254} and Markov decision processes \citep{Chowdhury_Zhou_2022, DBLP:journals/corr/abs-2010-07778}.
For a comprehensive overview of differential privacy and its application to other problems, see \citet{Dwork:2014:AFD:2693052.2693053}. 

The notion of using a sliding window mechanism (as we do in our proposed algorithm) to deal with a non-stationary environment has been employed in classical bandits \citep{GarivierSW} as well as Markov decision processes \citep{GajaneSW}.

\section{Non-Stationary Stochastic Corrupt Bandits}
A non-stationary stochastic corrupt bandits problem is formally characterized by a set of arms $A= \{1, \dots, K\}$ on which are indexed a list of unknown sub-Gaussian reward distributions $\{\RewardDist{a}{1} \}_{a\in A}, \dots, \{\RewardDist{a}{\numChanges_T} \}_{a\in A}$,
a list of unknown sub-Gaussian feedback distributions $\{\FeedbackDist{a}{1} \}_{a\in A}, \dots, \{\FeedbackDist{a}{\numChanges_T} \}_{a\in A}$, and a list of known \textit{mean-corruption functions} $\{g_a\}_{a\in A}$. Here, the total number of time steps (i.e., the \textit{horizon}) is indicated as $T$. The environment undergoes $\numChanges_T$ abrupt changes at unknown time steps called as \textit{breakpoints} and it remains constant in the intervals between two successive breakpoints.

For notational convenience, we assume that the first breakpoint occurs at $t=1$. From $i^{\text{th}}$ breakpoint till the subsequent breakpoint (or the horizon, in case of the last breakpoint), if the learner pulls an arm $ a\in A$ at time $t$, they receive a (hidden) reward $R_t$ drawn from the distribution $\RewardDist{a}{i}$ with mean $\MeanReAtIChanges{a}{i}$ and observe a feedback $F_t$ drawn from the distribution $\FeedbackDist{a}{i}$ with mean $\MeanFeAtIChanges{a}{i}$.
We assume that, for each arm, there exists a loose link between the reward and the feedback through a known \textit{corruption function} $g_a$ which maps the mean of the reward distribution to the mean of the feedback distribution :
$ g_a( \MeanReAtIChanges{a}{i} )= \MeanFeAtIChanges{a}{i}, \forall a\in A$ and $1\leq i \leq \numChanges_T$. Our proposed algorithm and the proven regret bound also work if the corruption function for an arm changes across time as long as the current corruption function is known. 

Note that these $g_a$ functions may be completely different from one arm to another.
For Bernoulli distributions, the reward distributions and the feedback distributions are in $[0,1]$ for all $a \in A$ and we assume all the corruption functions $\{g_a\}_{a \in A}$ to be continuous
in this interval. We also assume the corruption functions $\{g_a\}_{a \in A}$ to be strictly monotonic and denote the corresponding inverse functions by $g_a^{-1}$. The assumption of monotonicity is required for efficient learning as proved in \citet{pmlr-v83-gajane18a}.  

Another way to define the link between the reward and the feedback is to provide a \textit{corruption scheme} operator $\tilde{g}_a$ which maps the rewards into feedback distributions. 

\paragraph{Randomized Response}
Randomized response (a privacy protection technique introduced by \cite{Warner1965}) can also be simulated by a Bernoulli corrupt bandits problem and the corresponding corruption scheme $\tilde{g}_a$ is encoded as:
\begin{equation}
\label{eq:CorruptionMatrix}
  \mathds{M}_{a} \defined \kbordermatrix{%
      & 0  & 1  \\
    0 & p_{00}(a) & 1 - p_{11}(a) \\
    1 & 1 - p_{00}(a) & p_{11}(a) \\
  }
\end{equation}
Each item in $\mathds{M}_{a}$ denotes the probability of observing a particular feedback 
for a particular reward i.e.,
$
    \mathds{M}_a(y,x) \defined \mathds{P} \big( \text{Feedback from arm } a \IsEqual  y 
      \given
     \text{Reward from arm } a \IsEqual x \big).
$
The corresponding corruption function is $g_a(x) =  1 - p_{00}(a) + [p_{00}(a) + p_{11}(a) - 1]\cdot{} x.$

To measure the utility of an algorithm for this problem, we define the notion of regret in the following. Let us denote the mean reward of arm $a$ at time step $t$ as $\meanReward{a}{t}$. The objective of an algorithm, which chooses the arm $\ArmAt_{t}$ at time $t$ based only on the previously observed feedback, $F_1,\dots,F_{t-1}$, is to maximize the expected sum of rewards i.e., to achieve high utility. This is equivalent to minimizing the regret, 
$
\Regret(T) \defined \sum_{t=1}^{T} \optMean{t} - \bE\left[\sum_{t=1}^{T}\meanReward{\hat{a}_t}{t}\right],$
where $\optMean{t} \defined \max_{a \in A} \meanReward{a}{t}$.
Regret measures the performance of the algorithm against an omniscient policy that at each time step chooses the arm with the maximal mean reward. Thus, low regret translates to achieving high utility.

\section{The Proposed Algorithm}
To solve the problem at hand, we propose \SWKLUCBCF,
an adaptation of the \KLUCB \ algorithm of \citet{KLUCBJournal}. 
The algorithm takes as input: the window size $\windowsize$, a non-decreasing function $f$, the horizon $T$ and the corruptions functions $g_1, \dots, g_K$. We assume that the horizon T is known; an unknown $T$ can be handled using the doubling trick \citep{besson:hal-01736357}. We use $d(x,y)$ to denote the Kullback–Leibler divergence between two Bernoulli distributions with mean $x$ and $y$. 
We also use a shorthand of $x \wedge y$ to denote $\min(x,y)$.

At each time time step $t$, the algorithm computes an $\mathrm{Index}_a(t)$, which is an upper-confidence bound on $\meanReward{a}{t}$ built from a confidence interval on $\meanFeedback{a}{t}$ based on the KL-divergence. 
The quantity $N_a(t, \windowsize)$ denotes the number of times arm $a$ was chosen in the last $\windowsize$ time steps until time $t$.
Correspondingly, $\hat{\lambda}_a(t, \windowsize)$ denotes the empirical mean of the feedback observed from arm $a$ in the last $\windowsize$ time steps until time $t$: $\hat{\lambda}_a(t, w) \defined \frac{1}{N_a(t, \windowsize)}\sum_{s=\min\{1, t-\windowsize+1\}}^t F_s \cdot \ind_{(\ArmAt_s = a)}$.

Theorem~\ref{MainTheorem1}  gives an upper bound on the regret of \SWKLUCBCF. A more explicit bound is proved in the Appendix.

\begin{myTheorem}
\label{MainTheorem1} 
The regret of \SWKLUCBCF \ using  $f(x) \defined \log(x)+3\log(\log(x))$ and $\windowsize = \sqrt{ \frac{4eT}{\numChanges_T + 4}}$ on a  
Bernoulli non-stationary stochastic corrupt bandits problem with strictly monotonic and continuous corruption functions $\{g_a\}_{a \in A}$ at time $T$ is upper-bounded by \footnote{$\tilde{O}$ ignores logarithmic factors and constants.}
$$
\tilde{O}\left( \sum_{a \in A} \sqrt{\numChanges_T T} + \sum_{i=1}^{\numChanges_T} \sum_{a \neq \optArmAtIChanges{i}} \frac{\log{\left(  \sqrt{ \frac{T}{\numChanges_T}} \right)}}{d(\MeanFeAtIChanges{a}{i},g_a(\optMeanAtIChanges{i})} \right),
$$
where $\optArmAtIChanges{i}$ and $\optMeanAtIChanges{i}$ are the optimum arm and the corresponding optimal mean respectively after $i^{th}$ change and before the subsequent change.
\end{myTheorem}

The lower bound on regret in terms $T$ for classical non-stationary stochastic bandits is $\Omega(\sqrt{T})$ \citep{GarivierSW}. Theorem \ref{MainTheorem1} matches the lower bound up to logarithmic factors, so \SWKLUCBCF \ has near-optimal regret guarantees in terms of the time horizon $T$. The best known regret upper bounds for classical non-stationary stochastic bandits (e.g., \citet{pmlr-v99-auer19a}) also feature logarithmic terms besides the lower bound, hence our regret bound is in line with the best known results for analogous problems.
Moreover, the bound in Theorem \ref{MainTheorem1} also matches the best known regret bound in terms of $\numChanges_T$ for classical non-stationary stochastic bandits which is $O\sqrt{\numChanges_T}$. 

\begin{algorithm}[t]
{\caption{Sliding Window KLUCB \ for Non-Stationary Stochastic Corrupt Bandits  (\SWKLUCBCF)}}%
\label{algo:SWKLUCBCF}%
{
\begin{flushleft}
\textbf{Input:} 
Window size $\windowsize$,
a non-decreasing function $f:\mathds{N} \rightarrow \mathds{R}$, $T$,  monotonic and continuous corruption functions $g_1, \dots, g_K$ and $d(x,y) \defined \mathrm{KL}(\cB(x),\cB(y))$, 
\end{flushleft}
\begin{enumerate}
\item \textbf{Initialization:} Pull each arm once.
\item \textbf{for} time $t = K, \dots, T-1$ \textbf{do}
\begin{enumerate}
\item Compute for each arm $a \in A$ the quantity
\begin{align*}
    &
    \mathrm{Index}_a(t) \\
    & \defined  \max 
\left\{ q:\  N_a(t, \windowsize)\cdot{}d(\hat{\lambda}_a(t, \windowsize ), g_a(q)) \leq f\left( t \wedge \windowsize\right)\right\}
\end{align*}
\item Pull arm $\ArmAt_{t+1} \defined \operatornamewithlimits{argmax}\limits_{a \in A} {\operatorname{Index}_a(t)}$ and observe the feedback $F_{t+1}$.
\end{enumerate}
\textbf{end for}
\end{enumerate}
}
\end{algorithm}

We can use \SWKLUCBCF \ on non-stationary stochastic corrupts bandits where the corruption is done via randomized response. The following corollary bounds the resulting regret.
\begin{myCorollary}
\label{UB_corollary}
The regret of \SWKLUCBCF \ on a Bernoulli non-stationary stochastic corrupt bandits problem with randomized response using corruption matrices $\{\mathds{M}\}_{a \in A}$ at time $T$ is upper-bounded by
$$
\tilde{O}\left( \sum_{a \in A} \sqrt{\numChanges_T T} + \sum_{i=1}^{\numChanges_T} \sum_{a \neq \optArmAtIChanges{i}} \frac{\log{\left(  \sqrt{ \frac{T}{\numChanges_T}} \right)}}{(p_{00}(a) + p_{11}(a)-1)^2} \right).
$$
\end{myCorollary}
\noindent
This corollary follows from Theorem~\ref{MainTheorem1} and Pinsker's inequality: $d(x,y) > 2(x-y)^2$.
The term $(p_{00}(a) + p_{11}(a) -1)$ can be understood as the slope of the corruption function $g_a$.

\section{Corruption Mechanism to Preserve Local Privacy in Non-Stationary Environment}
\label{sec:privacy}
First, let us formally define local differential privacy.
\begin{myDefinition}
\label{def:diff_privacy}
(Locally differentially private mechanism) Any randomized mechanism $\mathcal{M}$ is $\epsilon$-locally differentially private for $\epsilon \geq 0$, if for all  $d_1, d_2 \in Domain(\mathcal{M})$ and for all $S \subset Range(\mathcal{M})$,
$$
\bP[\mathcal{M}(d_1) \in S] \leq e^\epsilon \cdot \bP [\mathcal{M}(d_2) \in S].
$$
\end{myDefinition}
As done in \citet{pmlr-v83-gajane18a}, a straightforward approach to achieve local differential privacy using corrupt bandits is to employ a corruption scheme on the user feedback.
This is similar to how randomized response is used in data collection by \citet{DBLP:conf/edbt/0009WH16}.    

\begin{myDefinition}
($\epsilon$-locally differentially private bandit feedback corruption scheme) A bandit feedback corruption scheme $\tilde{g}$ is $\epsilon$-locally differentially private for $\epsilon \geq 0$, if for all reward sequences $R_{t1}, \dots, R_{t2}$ and $R'_{t1} \dots, R'_{t2}$, and for all $\mathcal{S} \subset Range(\tilde{g})$, 
\begin{equation*} 
\bP [\tilde{g}(R_{t1}, \dots, R_{t2}) \in \mathcal{S} ] \leq e^\epsilon \cdot \bP [\tilde{g}(R'_{t1}, \dots, R'_{t2}) \in \mathcal{S} ].
\end{equation*}
\end{myDefinition}

When corruption is done by randomized response, 
local differential privacy requires that 
$\max_{ 1 \leq a \leq K}{ \Big( \frac{p_{00}(a)}{1 - p_{11}(a)}, \frac{p_{11}(a)}{1 - p_{00}(a)} \Big)} \leq e^{\epsilon} $.
%
%
From Corollary~\ref{UB_corollary}, we can see that to achieve lower regret, $p_{00}(a) + p_{11}$(a) is to be maximized for all $a \in A$. Using \citet[Result 1]{DBLP:conf/edbt/0009WH16}, we can state that, in order to achieve $\epsilon$-local differential privacy while maximizing $p_{00}(a) + p_{11}(a)$,
\begin{equation}
\label{eq:DPmatrix}
\mathds{M}_{a}=\kbordermatrix{%
      & 0  & 1  \\
    0 & \frac{e^{\epsilon}}{ 1 + e^\epsilon} & \frac{1}{ 1 + e^\epsilon } \\
    1 & \frac{1}{ 1 + e^\epsilon } & \frac{e^{\epsilon}}{ 1 + e^\epsilon} \\
  }.
\end{equation}
As it turns out, this is equivalent to the \textit{staircase} mechanism for local privacy
which is the optimal local differential privacy mechanism for low privacy regime \cite[Theorem 14]{JMLR:v17:15-135}. 
The trade-off between utility and privacy is controlled by $\epsilon$. 

\paragraph{}
Using the corruption parameters from Eq. (\ref{eq:DPmatrix}) with  Corollary~\ref{UB_corollary}, we arrive  at the following upper bound.
\begin{myCorollary}
\label{DPUB_corollary}At time $T$, 
the regret of \SWKLUCBCF \  with $\epsilon$-locally differentially private bandit feedback corruption scheme given by Eq. \eqref{eq:DPmatrix} is 
$
\tilde{O}\left( \sum_{a \in A} \sqrt{\numChanges_T T} + \sum_{i=1}^{\numChanges_T} \sum_{a \neq \optArmAtIChanges{i}} \frac{\log{\left(  \sqrt{ \frac{T}{\numChanges_T}} \right)}}{\left( \frac{e^\epsilon - 1}{e^\epsilon + 1}\right)^2} \right).
$
\end{myCorollary}
The term $\big( \frac{e^\epsilon - 1}{e^\epsilon + 1}\big)^2$ in the above expression conveys the relationship of the regret with the level of local differential privacy symbolized by $\epsilon$. For low values of $\epsilon$,  $ \big( \frac{e^\epsilon - 1}{e^\epsilon + 1}\big) \approx \epsilon / 2 $. This is in line with other bandit algorithms providing differential privacy (e.g., \citet{DBLP:conf/uai/MishraT15}).

\section{Elements of Mathematical Analysis}
Here, we provide a proof outline for Theorem~\ref{MainTheorem1}. Please refer to the Appendix for the complete proof.

We start by bounding the expected number of times a suboptimal arm (i.e., an arm other than the optimal arm at the time of selection) is pulled by the algorithm till horizon $T$. Recall that, at any time step $t$, \SWKLUCBCF \ pulls an arm maximizing an index defined as
\begin{align*}
    & \mathrm{Index}_a(t) \\
    &\defined \max \left\{ q:\  N_a(t, \windowsize)\cdot{}d\left(\hat{\lambda}_a(t, \windowsize), g_a(q)\right) \leq f\left(t \wedge \windowsize\right) \right\} \\
    &= \max g_a^{-1}\left( \left\{q: N_a(t, \windowsize)\cdot{}d\left(\hat{\lambda}_a(t, \windowsize), q\right) \leq f\left(t \wedge \windowsize\right)\right\} \right).
\end{align*}
%
We further decompose the computation of index as follows,
$$\mathrm{Index}_a(t) \defined
    \begin{cases}
      g_a^{-1}({\ell_a(t)}) & \text{if } g_a \text{ is decreasing}, \\
      g_a^{-1}({u_a(t)}) & \text{if } g_a \text{ is increasing}
    \end{cases}
$$
where,
\begin{align*}
\ell_a(t)  &\defined  \min \Big\{q: N_a(t, \windowsize)\cdot{}d\left(\hat{\lambda}_a(t, \windowsize), q\right) \leq f\left( t \wedge \windowsize \right)\Big\}, \\
u_a(t)  &\defined  \max \Big\{q: N_a(t, \windowsize)\cdot{}d\left(\hat{\lambda}_a(t, \windowsize), q\right) \leq f\left( t \wedge\windowsize \right) \Big\}.
\end{align*}

The interval $[\ell_a(t), u_a(t)]$ is a KL-based confidence interval on the mean feedback $\meanFeedback{a}{t}$ of arm $a$ at time $t$. 
This is in contrast to \KLUCB\ \citep{KLUCBJournal} where a confidence interval is placed on the mean reward. Furthermore, 
This differs from \KLUCBCF \ \cite{pmlr-v83-gajane18a} where the mean feedback of an arm remains the same for all the time steps and $f$ does not feature $\windowsize$.

In our analysis, we use the fact that when an arm $a$ is picked at time $t+1$ by \SWKLUCBCF, one of the following is true:
    Either the mean feedback of the optimal arm $\optArm{t}$ with mean reward $\optMean{t}$ is outside its confidence interval (i.e., $g_{\optArm{t}}(\optMean{t}) < \ell_{\optArm{t}}(t)$ or $g_{\optArm{t}}(\optMean{t}) > u_{\optArm{t}}(t)$) which is unlikely. Or, 
    the mean feedback of the optimal arm is where it should be, and then the fact that arm $a$ is selected indicates that the confidence interval on $\lambda_a$ cannot be too small as either $(u_a(t) \geq g_a(\optMean{t}))$ or $(\ell_a(t) \leq g_a(\optMean{t}))$. 
The previous statement follows from considering various cases depending on whether the corruption functions $g_a$ and $g_{\optArm{t}}$ are increasing or decreasing. 
We then need to control the two terms in the decomposition of the expected number of draws of arm $a$. The term regarding the ``unlikely" event, is bounded using the same technique as in the \KLUCB \ analysis, however with some added challenges due to the use of a sliding window. In particular, the analysis of a typical upper confidence bound algorithm for bandits relies on the fact that the confidence interval for any arm is always non-increasing, however this is not true while using a sliding window. To control the second term, depending on the monotonicity of the corruption functions $g_a$ and $g_{\optArm{t}}$, we need to meticulously adapt the arguments in \citet{KLUCBJournal} to control the number of draws of a suboptimal arm, as can be seen in the Appendix. 

\section{Concluding Remarks}
In this work, we proposed the setting of non-stationary stochastic corrupt bandits for preserving privacy while still maintaining high utility in sequential decision making in a changing environment. We devised an algorithm called \SWKLUCBCF \ and proved its regret upper bound which is near-optimal in the number of time steps and matches the best known bound for analogous problems in terms of the number of time steps and the number of changes. Moreover, we provided an optimal corruption scheme to be used with our algorithm in order to attain the dual goal of achieving high utility while maintaining the desired level of privacy. 

Interesting directions for future work include:
\begin{enumerate}
    \item Complete an empirical evaluation of the proposed algorithm on simulated as well as real-life data. 
    \item Characterize the changes in the environment by a variation budget (as done in \citet{NIPS2014_903ce922} for classical bandits) instead of the number of changes.
    \item Incorporate contextual information in the learning process. 
    \item Propose a Bayesian algorithm for non-stationary stochastic corrupt bandits.
    \item Propose a (near-)optimal differentially private algorithm which does not need to know the number of changes.
\end{enumerate}

\bibliography{main.bib}

\onecolumn

\appendix
\label{sec:Proof_MainTheorem1}
\section{Proof of Theorem \ref{MainTheorem1}}
\textit{Proof.} The proof follows along the lines of the proof for Theorem 2 from \citet{pmlr-v83-gajane18a}.

The index used by \SWKLUCBCF is defined by
\begin{align*}
    \mathrm{Index}_a(t) &\defined \max 
\left\{ q:\  N_a(t, \windowsize)\cdot{}d\left(\hat{\lambda}_a(t, \windowsize), g_a(q)\right) \leq f\left(t \wedge \windowsize\right) \right\} \\
&= \max g_a^{-1}\left( \left\{q: N_a(t, \windowsize)\cdot{}d\left(\hat{\lambda}_a(t, \windowsize), q\right) \leq f\left(t \wedge \windowsize\right)\right\} \right).
\end{align*}


\noindent
For the purpose of this proof, we further decompose the computation of index as follows,
$$\mathrm{Index}_a(t) \defined
    \begin{cases}
      g_a^{-1}({\ell_a(t)}) & \text{if } g_a \text{ is decreasing}, \\
      g_a^{-1}({u_a(t)}) & \text{if } g_a \text{ is increasing}
    \end{cases}
$$
where,
\begin{align*}
\ell_a(t)  &\defined  \min \left\{q: N_a(t, \windowsize)\cdot{}d\left(\hat{\lambda}_a(t, \windowsize), q\right) \leq f\left(  t \wedge \windowsize \right)\right\} \text{ and } \\
u_a(t)  &\defined  \max \left\{q: N_a(t, \windowsize)\cdot{}d\left(\hat{\lambda}_a(t, \windowsize), q\right) \leq f\left( t \wedge \windowsize \right) \right\}.
\end{align*}

Note that, the optimal arm at time $t$ is denoted as $\optArm{t}$ and $\optMean{t}$ is the corresponding optimal mean. Along the same lines, let $\ell_*(t) \defined \ell_{\optArm{t}}(t)$ and $u_*(t) \defined u_{\optArm{t}}(t)$. 

Let $N_a(t)$ be the number of times arm $a$ has been pulled till time $t$.
To get an upper bound on the regret of our algorithm, we first bound $\mathds{E}[N_a(t)]$ for all the non-optimal arms $a$ (i.e., $a \neq \optArm{t}$ at time $t$). Recall that $\meanReward{i}{t}$ is the mean reward of arm $i$ at time step $t$. Let us define $\noChange(\windowsize)$ as the set of indices $t \in \{K+1, \dots, T\}$ such that $\meanReward{i}{s} = \meanReward{i}{t}$ for all $ i \in \{1, \dots, K\}$ and all $t - \windowsize < s \leq t$. That is to say $\noChange(\windowsize)$ is the set of all time steps $t \in \{K+1, \dots, T\}$ for which there was no change in the previous $\windowsize$ time steps. 
Recall that $\ArmAt_{t}$ is the arm chosen by the algorithm at time step $t$. Then, 
\begin{align*}
\mathds{E}(N_a(T)) &= 1 + \sum_{t = K}^{T-1}\mathds{P}(\ArmAt_{t+1} \IsEqual a) \\
&\leq 1 + \numChanges_T \cdot \windowsize  + \sum_{K \leq t \leq T-1, \ t \in \noChange(\windowsize)} 
\bP(\ArmAt_{t+1} = a).
\end{align*}
Depending upon if $g_a$ and $g_{\optArm{t}}$ are increasing or decreasing there are four possible sub-cases:
\begin{itemize}
\item Both $g_{\optArm{t}}$ and $g_a$ are increasing. 
\begin{align*}
&(\ArmAt_{t+1} \IsEqual a) \\
&\subseteq \left(u_*(t) < g_{\optArm{t}}(\optMean{t})\right) \cup \left(\ArmAt_{t+1} \IsEqual a, u_*(t) \geq g_{\optArm{t}}(\optMean{t})\right) \\
&= \left(u_*(t) < g_{\optArm{t}}(\optMean{t})\right) \cup \left(\ArmAt_{t+1} \IsEqual a, g^{-1}_{\optArm{t}}(u_*(t)) \geq \optMean{t}\right) \qquad \text{since $g_{\optArm{t}}$ is increasing}\\
&= \left(u_*(t) < g_{\optArm{t}}(\optMean{t})\right) \cup \left(\ArmAt_{t+1} \IsEqual a, g^{-1}_a(u_a(t)) \geq \optMean{t}\right) \qquad \text{since $\operatorname{Index}_a \geq \operatorname{Index}_{\optArm{t}}$}\\
&= \left(u_*(t) < g_{\optArm{t}}(\optMean{t})\right) \cup \left(\ArmAt_{t+1} \IsEqual a, u_a(t) \geq g_a(\optMean{t})\right) \qquad \text{since $g_a$ is increasing.}
\end{align*}
\begin{align}
\label{exp_of_N_eq1}
\therefore \mathds{E}(N_a(T)) \leq &1 + \numChanges_T \cdot \windowsize +  \sum_{K \leq t \leq T-1, \ t \in \noChange(\windowsize)} \mathds{P}\left(u_*(t) < g_{\optArm{t}}(\optMean{t})\right) \nonumber \\
&+ \sum_{K \leq t \leq T-1, \ t \in \noChange(\windowsize)} \mathds{P}\left(\ArmAt_{t+1} \IsEqual a, u_a(t) \geq g_a(\optMean{t})\right).
\end{align}
\item $g_{\optArm{t}}$ is decreasing and $g_a$ is increasing. 
\begin{align*}
&(\ArmAt_{t+1} \IsEqual a) \\
&\subseteq \left(\ell_*(t) > g_{\optArm{t}}(\optMean{t})\right) \cup \left(\ArmAt_{t+1} \IsEqual a, \ell_*(t) \leq g_{\optArm{t}}(\optMean{t})\right) \\
&= \left(\ell_*(t) > g_{\optArm{t}}(\optMean{t})\right) \cup \left(\ArmAt_{t+1} \IsEqual a, g_{\optArm{t}}^{-1}(\ell_*(t)) \geq \optMean{t}\right) \qquad \text{since $g_{\optArm{t}}$ is decreasing} \\
&= \left(\ell_*(t) > g_{\optArm{t}}(\optMean{t})\right) \cup \left(\ArmAt_{t+1} \IsEqual a, g_a^{-1}(u_a(t)) \geq \optMean{t}\right) \qquad \text{since $\operatorname{Index}_a \geq \operatorname{Index}_{\optArm{t}}$}\\
&= \left(\ell_*(t) > g_{\optArm{t}}(\optMean{t})\right) \cup \left(\ArmAt_{t+1} \IsEqual a, u_a(t) \geq g_a(\optMean{t})\right) \qquad \text{since $g_a$ is increasing.}
\end{align*}
\begin{align}
\label{exp_of_N_eq2}
\therefore \mathds{E}(N_a(T)) \leq &1 + \numChanges_T \cdot \windowsize +  \sum_{K \leq t \leq T-1, \ t \in \noChange(\windowsize)} \mathds{P}\left(\ell_*(t) > g_{\optArm{t}}(\optMean{t})\right) \nonumber \\
&+ \sum_{K \leq t \leq T-1, \ t \in \noChange(\windowsize)} \mathds{P}\left(\ArmAt_{t+1} \IsEqual a, u_a(t) \geq g_a(\optMean{t})\right).
\end{align}
\item $g_{\optArm{t}}$ is increasing and $g_a$ is decreasing. 
\begin{align*}
&(\ArmAt_{t+1} \IsEqual a) \\
&\subseteq \left(u_*(t) < g_{\optArm{t}}(\optMean{t})\right) \cup \left(\ArmAt_{t+1} \IsEqual a, u_*(t) \geq g_{\optArm{t}}(\optMean{t})\right) \\
&= \left(u_*(t) < g_{\optArm{t}}(\optMean{t})\right) \cup \left(\ArmAt_{t+1} \IsEqual a, g_{\optArm{t}}^{-1}(u_*(t)) \geq \optMean{t}\right) \qquad \text{since $g_{\optArm{t}}$ is increasing}\\ 
&= \left(u_*(t) < g_{\optArm{t}}(\optMean{t})\right) \cup \left(\ArmAt_{t+1} \IsEqual a, g_a^{-1}(\ell_a(t)) \geq \optMean{t}\right) \qquad \text{since $\operatorname{Index}_a>\operatorname{Index}_{\optArm{t}}$}\\ 
&= \left(u_*(t) < g_{\optArm{t}}(\optMean{t})\right) \cup \left(\ArmAt_{t+1} \IsEqual a, \ell_a(t) \leq g_a(\optMean{t})\right) \qquad \text{since $g_a$ is decreasing.}
\end{align*}
\begin{align}
\label{exp_of_N_eq3}
\therefore \mathds{E}(N_a(T)) \leq &1 + \numChanges_T \cdot \windowsize +  \sum_{K \leq t \leq T-1, \ t \in \noChange(\windowsize)} \mathds{P}\left(u_*(t) < g_{\optArm{t}}(\optMean{t})\right) \nonumber \\
&+ \sum_{K \leq t \leq T-1, \ t \in \noChange(\windowsize)} \mathds{P}\left(\ArmAt_{t+1} \IsEqual a, \ell_a(t) \leq g_a(\optMean{t})\right).
\end{align}

\item $g_{\optArm{t}}$ is decreasing and $g_a$ is decreasing.
\begin{align*}
&(\ArmAt_{t+1} \IsEqual a) \\
&\subseteq \left(\ell_*(t) > g_{\optArm{t}}(\mu_{\optArm{t}})\right) \cup \left(\ArmAt_{t+1} \IsEqual a, \ell_*(t) \leq g_{\optArm{t}}(\mu_{\optArm{t}}\right) \\
&= \left(\ell_*(t) > g_{\optArm{t}}(\mu_{\optArm{t}})\right) \cup \left(\ArmAt_{t+1} \IsEqual a, g_{\optArm{t}}^{-1}(\ell_*(t)) \geq \mu_{\optArm{t}}\right) \qquad \text{since $g_{\optArm{t}}$ is decreasing} \\
&= \left(\ell_*(t) > g_{\optArm{t}}(\mu_{\optArm{t}})\right) \cup \left(\ArmAt_{t+1} \IsEqual a, g_a^{-1}(\ell_a(t)) \geq \mu_{\optArm{t}}\right) \qquad \text{since $\operatorname{Index}_a>\operatorname{Index}_{\optArm{t}}$} \\
&= \left(\ell_*(t) > g_{\optArm{t}}(\mu_{\optArm{t}})\right) \cup \left(\ArmAt_{t+1} \IsEqual a, \ell_a(t) \leq g_a(\mu_{\optArm{t}})\right) \qquad \text{since $g_a$ is decreasing.}
\end{align*}
\begin{align}
\label{exp_of_N_eq4}
\therefore \mathds{E}(N_a(T)) \leq &1 + \numChanges_T \cdot \windowsize +  \sum_{K \leq t \leq T-1, \ t \in \noChange(\windowsize)} \mathds{P}\left(\ell_*(t) > g_{\optArm{t}}(\mu_{\optArm{t}})\right) \nonumber \\
&+ \sum_{K \leq t \leq T-1, \ t \in \noChange(\windowsize)} \mathds{P}\left(\ArmAt_{t+1} \IsEqual a, \ell_a(t) \leq g_a(\mu_{\optArm{t}})\right).
\end{align}
\end{itemize}

We first upper bound the two sums 
\begin{equation}\sum_{K \leq t \leq T-1, \ t \in \noChange(\windowsize)} \mathds{P}\left(u_*(t) < g_{\optArm{t}}(\optMean{t})\right) \ \ \text{and} \ \ \sum_{K \leq t \leq T-1, \ t \in \noChange(\windowsize)} \mathds{P}\left(\ell_*(t) > g_{\optArm{t}}(\mu_{\optArm{t}})\right)\label{eq:FirstTerms}\end{equation}
using that  $\ell_*(t)$ and $u_*(t)$ are respectively lower and upper confidence bound on $g_{\optArm{t}}(\optMean{t})$. Recall that $\min\left\{ t, \windowsize\right\}$ is denoted as $t \wedge \windowsize$. 
\begin{align}
& \bP\left(u_{\optArm{t}} < g_{\optArm{t}}(\optMean{t})\right) \nonumber \\
& \leq  \bP\left(g_{\optArm{t}}(\optMean{t}) > \hat{\lambda}_{\optArm{t}}(t, \windowsize) \text{ and } N_{\optArm{t}}(t, \windowsize) \cdot  d\left(\hat{\lambda}_{\optArm{t}}(t, \windowsize), g_{\optArm{t}}(\optMean{t})\right) \geq f\left( t \wedge \windowsize\right)\right) \nonumber \\
& \leq \bP\left(\exists s \in \{1,\dots, (t \wedge \windowsize)\} :  g_{\optArm{t}}(\optMean{t}) > \hat{\lambda}_{\optArm{t},s} \text{ and } s \cdot d(\hat{\lambda}_{\optArm{t},s}, g_{\optArm{t}}(\optMean{t})) \geq f\left( t \wedge \windowsize\right)\right) \nonumber \\
& \leq  min\left\{ 1, e \left \lceil f\left( t \wedge \windowsize\right) \log{t} \right \rceil e^{-f\left( t \wedge \windowsize\right)} \right\} \label{eq:FirstTwoUpperBoundA},
\end{align}
where the upper bound follows from Lemma 2 in \citet{KLUCBJournal}, and the fact that $\hat{\lambda}_{\optArm{t},s}$ is the empirical mean of $s$ Bernoulli samples with mean $g_{\optArm{t}}(\optMean{t})$.
Similarly, one has 
\begin{equation}
\mathds{P}\left(\ell_*(t) > g_{\optArm{t}}(\mu_{\optArm{t}})\right)  \leq  min\left\{ 1, e \left \lceil f\left( t \wedge \windowsize\right) \log{t} \right \rceil e^{-f\left( t \wedge \windowsize\right)} \right\}. \label{eq:FirstTwoUpperBoundB}
\end{equation}
As $f(x) \defined \log{x} + 3(\log{\log{x}})$, for $x\geq3$,
\[e \lceil f(x) \log{x} \rceil \leq 4e \log^2{x}.\]
Then, using Eq. \eqref{eq:FirstTwoUpperBoundA} and Eq. \eqref{eq:FirstTwoUpperBoundB}, the two quantities in Eq. \eqref{eq:FirstTerms} can be upper bounded by 
\begin{align*}
1 + \sum_{t = 3}^{T-1} e \left \lceil f\left( t \wedge \windowsize \right) \log{t} \right \rceil e^{-f\left( t \wedge \windowsize\right)} & \leq 
 1 + \sum_{t = 3}^{T-1} 4e \cdot \log^2 \left({t \wedge \windowsize}\right) \cdot e^{-f(t \wedge \windowsize)} \nonumber \\
&= 1 + 4e\sum_{t = 3}^{T-1} \frac{1}{(t \wedge \windowsize) \cdot  \log{(t \wedge \windowsize)}} \nonumber \\
&= 1 + 4e\sum_{t = 3}^{\windowsize} \frac{1}{(t \wedge \windowsize) \cdot  \log{(t \wedge \windowsize)}}  \ + \ 4e\sum_{t = \windowsize+1}^{T} \frac{1}{(t \wedge \windowsize) \cdot  \log{(t \wedge \windowsize)}}  \nonumber \\
&\leq 1 + 4e\sum_{t = 3}^{\windowsize} \frac{1}{ 3 \log{3}} + 4e\sum_{t = \windowsize+1}^{T} \frac{1}{ \windowsize \log{\windowsize}}  \nonumber \\
&\leq 1 + \frac{4 e \windowsize}{3 \log{3}} + \frac{4 e T}{\windowsize \log{\windowsize}}.
\end{align*}
This proves that 
\begin{align}
\sum_{K \leq t \leq T-1, \ t \in \noChange(\windowsize)} \mathds{P}\left(u_*(t) < g_{\optArm{t}}(\optMean{t})\right) &\leq  1 + \frac{4 e \windowsize}{3 \log{3}} + \frac{4 e T}{\windowsize \log{\windowsize}} \quad \text{and},\label{term1a} \\
\sum_{K \leq t \leq T-1, \ t \in \noChange(\windowsize)} \mathds{P}\left(\ell_*(t) > g_{\optArm{t}}(\mu_{\optArm{t}})\right) &\leq 1 + \frac{4 e \windowsize}{3 \log{3}} + \frac{4 e T}{\windowsize \log{\windowsize}} \label{term1b}.
\end{align}

We now turn our attention to the other two sums involved in the upper bound we gave for $\mathds{E}(N_a(T))$. Let the unknown time-step at which $i^{th}$ change occurs be denoted as $\timeOfChange{i}$. For notational convenience, we assume that the first change occurs at $t=1$ so $\timeOfChange{1} = 1$ and change $L+1$  takes place at $t= T+1$ where $T$ is the horizon.
We introduce the notation $d^+(x,y) = d(x,y) \cdot \ind_{(x<y)}$ and $d^-(x,y) = d(x,y) \cdot \ind_{(x>y)}$. So we can write, when $g_a$ is increasing,
\begin{align*}
 & \sum_{K \leq t \leq T-1, \ t \in \noChange(\windowsize)} \mathds{P}\left(\ArmAt_{t+1} \IsEqual a, u_a(t) \geq g_a(\optMean{t})\right) \\
&\leq  \sum_{i=1}^{L} \sum_{\timeOfChange{i}  \leq t  < \timeOfChange{i+1}-1, \  t \in \noChange(\windowsize)}
 \mathds{P}\left(\ArmAt_{t+1} \IsEqual a, u_a(t) \geq g_a(\optMean{t})\right) \\ 
 &= \bE\left[\sum_{i=1}^{L} \sum_{\timeOfChange{i}  \leq t  < \timeOfChange{i+1}-1, \  t \in \noChange(\windowsize)} \ind_{\ArmAt_{t+1} \IsEqual a} \cdot \ind_{N_a(t, w) \cdot d^+(\hat{\lambda}_{a,N_a(t, w)},g_a(\optMean{t})) \leq f(t \wedge \windowsize )}\right]  \\
 &\leq \bE\left[ \sum_{i=1}^{L} \sum_{\timeOfChange{i}  \leq t  < \timeOfChange{i+1}-1, \  t \in \noChange(\windowsize)} \sum_{s=1}^{t \wedge \windowsize} \ind_{\ArmAt_{t+1} \IsEqual a} \cdot \ind_{N_a(t, \windowsize) \IsEqual s} \cdot \ind_{s \cdot d^+(\hat{\lambda}_{a,s},g_a(\optMean{t})) \leq f(t \wedge \windowsize )} \right] \\
&\leq \bE\left[ \sum_{i=1}^{L} \sum_{\timeOfChange{i}  \leq t  < \timeOfChange{i+1}-1, \  t \in \noChange(\windowsize)} \sum_{s=1}^{t \wedge \windowsize} \ind_{\ArmAt_{t+1} \IsEqual a} \cdot \ind_{N_a(t) \IsEqual s} \cdot \ind_{s \cdot d^+(\hat{\lambda}_{a,s},g_a(\optMean{t})) \leq f(t \wedge \windowsize )} \right] \\
 &\leq \bE\Big[ \sum_{i=1}^{L}  \sum_{s=1}^{t \wedge \windowsize}  \ind_{s \cdot d^+(\hat{\lambda}_{a,s},g_a(\optMean{t})) \leq f(t \wedge \windowsize )} \underbrace{\sum_{\timeOfChange{i}  \leq t  < \timeOfChange{i+1}-1, \  t \in \noChange(\windowsize)} \ind_{\ArmAt_{t+1} \IsEqual a} \cdot \ind_{N_a(t) \IsEqual s}}_{\leq 1}   \Big]. \end{align*}
In the above, the penultimate steps follows from the fact that the event $N_a(t, \windowsize) = s$ is subsumed by the event $N_a(t) = s$.
So, one obtains, when $g_a$ is increasing, 
\begin{equation} 
\sum_{K \leq t \leq T-1, \ t \in \noChange(\windowsize)} \mathds{P}\left(\ArmAt_{t+1} \IsEqual a, u_a(t) \geq g_a(\optMean{t})\right) \leq \bP \left( \sum_{l=1}^{L}  \sum_{s=1}^{t \wedge \windowsize}  s \cdot d^+(\hat{\lambda}_{a,s},g_a(\optMean{t})) \leq f(t \wedge \windowsize ) \right).
\label{eq:SecondTermIncreasing}\end{equation}
Using similar arguments, one can show that when $g_a$ is decreasing,
\begin{equation}
\sum_{K \leq t \leq T-1, \ t \in \noChange(\windowsize)} \mathds{P}\left(\ArmAt_{t+1} \IsEqual a, \ell_a(t) \leq g_a(\mu_{\optArm{t}})\right) \leq \bP \left( \sum_{l=1}^{L}  \sum_{s=1}^{t \wedge \windowsize}  s \cdot d^-(\hat{\lambda}_{a,s},g_a(\optMean{t})) \leq f(t \wedge \windowsize ) \right).
\label{eq:SecondTermDecreasing}\end{equation}

Recall that $\MeanReAtIChanges{a}{i}$ is the mean reward of arm $a$ after $i^{th}$ change and before the subsequent change. Correspondingly, let $\MeanFeAtIChanges{a}{i}$ be the mean feedback of arm $a$ after $i^{th}$ change and and before the subsequent change. Furthermore, let $\optMeanAtIChanges{i}$ be the optimum mean after $i^{th}$ change and and before the subsequent change. 

Using Appendix A.2. of \cite{KLUCBJournal}, the quantity in the right-hand side of \eqref{eq:SecondTermIncreasing} can be upper-bounded by 
\begin{equation}  \sum_{i=1}^{L} \frac{ f(\windowsize)}{d(\MeanFeAtIChanges{a}{i},g_a(\optMeanAtIChanges{i})}
+ \sum_{i=1}^{L} \sqrt{2\pi}\sqrt{\frac{d'(\MeanFeAtIChanges{a}{i},g_a(\optMeanAtIChanges{i})^2}{(d(\MeanFeAtIChanges{a}{i},g_a(\optMeanAtIChanges{i})^3}}\sqrt{f(\windowsize)} + \sum_{i=1}^{L} 2 \left(\frac{d'(\MeanFeAtIChanges{a}{i},g_a(\optMeanAtIChanges{i})}{d(\MeanFeAtIChanges{a}{i},g_a(\optMeanAtIChanges{i})}\right)^{2} +1 . \label{eq:SecondTermIncreasingFinal}\end{equation}
For \eqref{eq:SecondTermDecreasing}, noting that $d^-(x,y) = d^+(1-x,1-y)$, one has 
\begin{align*}
\bP \left(  s \cdot d^-(\hat{\lambda}_{a,s},g_a(\optMean{t})) \leq  f(t \wedge \windowsize )\right) =& \bP\left( s \cdot d^+(1-\hat{\lambda}_{a,s},1-g_a(\optMean{t})) \leq f(t \wedge \windowsize )  \right)\\
=& \bP\left( s \cdot d^+(\hat{\mu}_{a,s},1-g_a(\optMean{t})) \leq f(t \wedge \windowsize )  \right),
\end{align*}
where $\hat{\mu}_{a,s} \defined 1-\hat{\lambda}_{a,s}$, is the empirical mean of $s$ observations of a Bernoulli random variable with mean $1-\lambda_a < 1 - g_a(\optMean{t})$. Hence, the analysis of \cite{KLUCBJournal} can be applied, and using that $d(1-x,1-y)=d(x,y)$ and $d'(1-x,1-y)=-d'(x,y)$, the right hand side of \eqref{eq:SecondTermDecreasing} can also be upper bound by \eqref{eq:SecondTermIncreasingFinal}.

Combining inequalities \eqref{term1a}, \eqref{term1b} and \eqref{eq:SecondTermIncreasing},\eqref{eq:SecondTermDecreasing}, \eqref{eq:SecondTermIncreasingFinal} with the initial decomposition of $\bE[N_a(T)]$, and substituting $f(x) \defined \log(x) + 3\log\log(x)$ yields in all cases, 
\begin{align}
\bE[N_a(T)] 
\leq 
& \  \numChanges_T \cdot \windowsize + \frac{4 e \windowsize}{3 \log{3}} + \frac{4 e T}{\windowsize \log{\windowsize}} +  \sum_{i=1}^{\numChanges_T} \frac{ f(\windowsize) }{d(\MeanFeAtIChanges{a}{i},g_a(\optMeanAtIChanges{i})} \nonumber \\
& \qquad + \sum_{i=1}^{\numChanges_T} \sqrt{2\pi}\sqrt{\frac{d'(\MeanFeAtIChanges{a}{i},g_a(\optMeanAtIChanges{i})^2}{(d(\MeanFeAtIChanges{a}{i},g_a(\optMeanAtIChanges{i})^3}}\sqrt{f(\windowsize)} \nonumber \\
& \qquad + \sum_{i=1}^{\numChanges_T} 2 \left(\frac{d'(\MeanFeAtIChanges{a}{i},g_a(\optMeanAtIChanges{i})}{d(\MeanFeAtIChanges{a}{i},g_a(\optMeanAtIChanges{i})}\right)^{2} + 5 \nonumber \\
&\leq (\numChanges_T + 4 )\cdot \windowsize + \frac{4 e T}{\windowsize \log{\windowsize}} +  \sum_{i=1}^{\numChanges_T} \frac{ \log(\windowsize) + 3\log{\log(\windowsize)}}{d(\MeanFeAtIChanges{a}{i},g_a(\optMeanAtIChanges{i})} \nonumber \\
& \qquad + \sum_{i=1}^{\numChanges_T} \sqrt{2\pi}\sqrt{\frac{d'(\MeanFeAtIChanges{a}{i},g_a(\optMeanAtIChanges{i})^2}{(d(\MeanFeAtIChanges{a}{i},g_a(\optMeanAtIChanges{i})^3}}\sqrt{ \log(\windowsize) + 3\log{\log(\windowsize)}} \nonumber \\
& \qquad + \sum_{i=1}^{\numChanges_T} 2 \left(\frac{d'(\MeanFeAtIChanges{a}{i},g_a(\optMeanAtIChanges{i})}{d(\MeanFeAtIChanges{a}{i},g_a(\optMeanAtIChanges{i})}\right)^{2} +  5. \label{eq:BoundNumberOfPlays}
\end{align}

\noindent Minimizing the leading terms in the RHS from eq. \eqref{eq:BoundNumberOfPlays} via taking the first derivative with respect to $\windowsize$ and equating it to $0$, leads to solving for $w$ in 
\begin{align*}
    & \ \ \frac{\windowsize^2 \left( \log^2{\windowsize} \right)}{\log{\windowsize} + 1} = \frac{4e T}{\numChanges_T + 4} \\
    & \simeq \ \windowsize^2 \log{(\windowsize^2)} = \frac{8 e T}{\numChanges_T + 4 } 
\end{align*}
Here, $\windowsize$ must be positive for the log to exist, so we can write $\windowsize^2 = e^u$ for some $u$, and the equation becomes
\begin{equation*}
    u e^u = \frac{8 e T}{\numChanges_T + 4}.
\end{equation*}
This equation has no solution in an elementary expression, although it can be expressed in terms of the Lambert W function \citep{Corless1996}. Opting for an elementary expression for $\windowsize$, we can choose $\windowsize = \sqrt{ \frac{4eT}{\numChanges_T + 4}}$, which leads to the following bound, 
\begin{align*}
    \bE[N_a(T)]  &\leq \sqrt{4 e (\numChanges_T + 4) T} + \frac{\sqrt{4 e (\numChanges_T + 4) T}}{\log{\left(\sqrt{ \frac{4eT}{ \numChanges_T + 4}}\right)}} + \sum_{i=1}^{\numChanges_T} \frac{ \log{\left(  \sqrt{ \frac{4eT}{\numChanges_T + 4}} \right)} + 3\log{\log{\left(  \sqrt{ \frac{4eT}{\numChanges_T + 4}} \right)}}}{d(\MeanFeAtIChanges{a}{i},g_a(\optMeanAtIChanges{i})} \\
    & \ + \sum_{i=1}^{\numChanges_T} \sqrt{2\pi}\sqrt{\frac{d'(\MeanFeAtIChanges{a}{i},g_a(\optMeanAtIChanges{i})^2}{(d(\MeanFeAtIChanges{a}{i},g_a(\optMeanAtIChanges{i})^3}}\sqrt{ \log{\left(  \sqrt{ \frac{4eT}{\numChanges_T + 4}} \right)} + 3\log{\log{\left(  \sqrt{ \frac{4eT}{\numChanges_T + 4}} \right)}}} \nonumber \\
    & \ + \sum_{i=1}^{\numChanges_T} 2 \left(\frac{d'(\MeanFeAtIChanges{a}{i},g_a(\optMeanAtIChanges{i})}{d(\MeanFeAtIChanges{a}{i},g_a(\optMeanAtIChanges{i})}\right)^{2} +  5.
\end{align*}
Since the rewards are bounded in $[0,1]$ for Bernoulli non-stationary stochastic bandits, the regret is upper-bounded by,
\begin{equation*}
     \tilde{O}\left( \sum_{a \in A} \sqrt{\numChanges_T T} + \sum_{a \neq \optArmAtIChanges{i}} \sum_{i=1}^{\numChanges_T} \frac{\log{\left(  \sqrt{ \frac{T}{\numChanges_T}} \right)}}{d(\MeanFeAtIChanges{a}{i},g_a(\optMeanAtIChanges{i})}   \right).
\end{equation*}
Assuming that $\numChanges_T = \left( T^\beta \right)$ for some $\beta \in [0,1)$, the expected regret is upper bounded as $\tilde{O}\left(T^{(1 + \beta)/2}\right)$. In particular, if $\beta = 0$, the number of breakpoints is upper-bounded by $\numChanges$ independently of $T$, then with $\windowsize = \sqrt{ \frac{4eT}{\numChanges + 4}}$, the upper bound is $\tilde{O}\left( \sqrt{\numChanges T} \right)$.

\end{document}